# A Fuzzy Logic Approach to Target Tracking

Chin-Wang Tao and Wiley E. Thompson
Electrical and Computer Engineering Dept.
New Mexico State University
Las Cruces, NM 88003

## Abstract

This paper discusses a target tracking problem in which no dynamic mathematical model is explicitly assumed. A nonlinear filter based on the fuzzy If-then rules is developed. A comparison with a Kalman filter is made, and empirical results show that the performance of the fuzzy filter is better. Intensive simulations suggest that theoretical justification of the empirical results is possible.

## I. INTRODUCTION

Target tracking is a very important problem in many areas, including military, space, and industrial applications. Because of measurement noise, target position measurements are often uncertain. As is well known, when the uncertainty may be represented statistically and the system dynamics are available, the Kalman filter has been very successful, according to its optimal property. However, when the system dynamics are unknown or the parameter variation is not negligible or the statistical model can not be properly applied to the uncertainty involved, the performance of the Kalman filter degrades, and alternative methods for tracking are necessary. An example described in a paper (Schlee 1967) shows the effect of erroneous system dynamics of a target on a Kalman filter. Specifically, if a Kalman filter is first designed for tracking a target which is assumed to have a constant position, then the Kalman filter diverges when the target moves at constant velocity. Moreover, the erroneous uncertainty model might cause divergence in a Kalman filter (Heffes 1966). In many real world problems, the knowledge of the moving target dynamics and the statistical uncertainty model are not available. In situations where the knowledge for constructing a mathematical model for the control process is not available, the use of expert knowledge is a reasonable alternative. Since expert knowledge is expressed in terms of linguistic rules, the use of fuzzy logic (Zadeh, 1965) is appropriate. Therefore, based on the idea of applying fuzzy logic in expert systems, where If-then rules represent the knowledge from experts, a fuzzy approach to tracking estimators using fuzzy If-then rules is a promising research area (Huang 1990). In this paper, a fuzzy filter is designed with only one linguistic variable needed for the fuzzy tracking estimator in order to reduce the complexity, and a bell-shaped function is chosen for the membership function of a fuzzy set.

The motivation and design of the filter based on fuzzy logic are discussed in sections II and III. Simulations in section IV present impressive performance and comparisons with a Kalman filter as in (Kosko 1992).

## II. PROBLEM FORMULATION

The problem considered here is that of target tracking based on measurements taken in a noisy environment. It is assumed that the exact dynamic model of the moving target is unknown and that only position measurements are available at each time index k, denoted as

{ x(k) : k=1,2,...,n; where n is finite }

A fuzzy tracking filter based on an inherent property of a large class of targets is designed for target tracking in order to avoid the difficulties in constructing a mathematical model used by a Kalman filter and in order to avoid the degradation of the performance of a Kalman filter based upon an inappropriate model. The basic problems involved here are the fuzzy partitions of the universe of discourse of the variables, the derivation of the fuzzy If-then rules, and defuzzification. Each of the problems are discussed in section III.

## III. DESIGN OF THE FUZZY TRACKER

A fuzzy filter, including a fuzzy estimator, is shown in Figure 1. The problem considered here is the construction of a fuzzy filter for moving targets with smooth trajectories. Specifically, it is assumed that a moving target can not change its direction very quickly, i.e., the difference $\theta(k)$ of the angles $\theta_1(k)$ and $\theta_1(k-1)$ shown in the following equations can not be very large:

$$\theta_1(k) = \tan^{-1}((x(k)-x_f(k-1))/T)$$

$$\theta_1(k-1) = \tan^{-1}((x_f(k-1)-x_f(k-2))/T)$$

T = sampling period;    $\theta(k) = \theta_1(k) - \theta_1(k-1)$

where k needs to be greater than 2. The input position variable x(k), k>2, is then transformed into an angle variable $\theta(k)$ in order to utilize the smoothness assumption. The output position variable $x_f(k)$ of the fuzzy filter is assumed to be equal to the input position variable x(k) when k is less than or equal to 2.

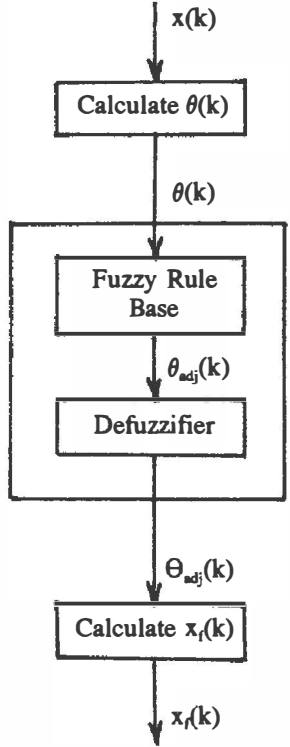

Figure 1: Fuzzy Filter

The smoothness assumption described above formulates the fuzzy If-then rules for the fuzzy controller in the fuzzy filter, for example:

R1: If the angle difference ($\theta(k)$) is positive small, then the angle adjustment ($\theta_{adj}(k)$) is negative small.

The "positive small" and "negative small" are not exact regions but are fuzzy sets which are defined by fuzzy membership functions (part A). For an input variable $\theta(k)$, it is necessary to find the corresponding degree of "positive small" for the variables $\theta(k)$ to be able to apply the fuzzy If-then rule R1. By applying and combining the fuzzy If-then rules, a fuzzy set "appropriate control" is obtained. For each particular input variable $\theta(k)$, every possible value of $\theta_{adj}(k)$ in its universe of discourse gives a degree of appropriateness for the control. In order to obtain the best $\Theta_{adj}(k)$, which is a crisp value, to be the output of the fuzzy controller, the defuzzification technique is needed. Since the best $\Theta_{adj}(k)$ is obtained, the algorithm below is used to obtain the filtered target position.

$$x_f(k) = \tan(\theta(k) + \Theta_{adj}(k) + \theta_1(k-1))T + x_f(k-1); \quad k>2$$

The following parts of this section give the details in the design of this fuzzy controller from a fuzzy partition of the universe of discourse of variables ($\theta(k), \theta_{adj}(k)$) to defuzzification.

### A. FUZZY PARTITION AND MEMBERSHIP FUNCTIONS

First, the universe of discourse of the variables $\theta(k)$ and $\theta_{adj}(k)$ needs to be defined. The closed range $[-\pi,\pi]$ is considered to be a universe of discourse $U_1$ for the input variable $\theta(k)$. In order to create effective and stable (with respect to parameter $\sigma$) membership functions, a mapping function $f_i: U_1 \rightarrow U$, specifically,

$$y_i(k) = f_i(\theta(k)) = (\theta(k)/\gamma)a$$

where $a = 2\sigma(\sqrt{2\log 2})$, $\gamma = 30$

is applied on the universe of discourse $U_1$ which maps $U_1$ to a universe of discourse $U=[-6a,6a]$. Secondly, the U is fuzzily partitioned into a collection of fuzzy sets corresponding to those fuzzy sets in the fuzzy If-then rules. Each of these fuzzy sets is specified by the names and the support of each fuzzy set as indicated in Table 1. These letters p, n, ze, s, m, b, v in the names of the fuzzy sets correspond, respectively, to the meaning of positive, negative, zero, small, big, and very. For example, pvvb means "positive very very big". Actually, the name of each fuzzy set in Table 1 really means that the element with membership value 1 in this fuzzy set has the property which is described by the name of this fuzzy set.

By changing the $\gamma$ in the mapping function $f_i$ to be 10, a mapping function $f_o$ is formed to map the universe of discourse $U_2=[-\pi/3,\pi/3]$ of the output variable $\theta_{adj}(k)$ to the same universe of discourse U, and the fuzzy partition is the same as in Table 1. That is,

$$y_o(k) = f_o(\theta_{adj}(k)) = (\theta_{adj}(k)/\gamma)a, \quad \text{where } \gamma = 10$$

Table 1: Fuzzy Sets and Supports

| Support of the Fuzzy Set | Fuzzy Set |
|---|---|
| $5a < y_i(k)(y_o(k)) \leq 6a$ | pvvb |
| $4a < y_i(k)(y_o(k)) < 6a$ | pvb |
| $3a < y_i(k)(y_o(k)) < 5a$ | pb |
| $2a < y_i(k)(y_o(k)) < 4a$ | pm |
| $a < y_i(k)(y_o(k)) < 3a$ | ps |
| $0 < y_i(k)(y_o(k)) < 2a$ | pv |
| $-a < y_i(k)(y_o(k)) < a$ | ze |
| $-2a < y_i(k)(y_o(k)) < 0$ | nvs |
| $-3a < y_i(k)(y_o(k)) < -a$ | ns |
| $-4a < y_i(k)(y_o(k)) < -2a$ | nm |
| $-5a < y_i(k)(y_o(k)) < -3a$ | nb |
| $-6a < y_i(k)(y_o(k)) < -4a$ | nvb |
| $-6a \leq y_i(k)(y_o(k)) < -5a$ | nvvb |

The membership function of each fuzzy set is defined by the bell-shaped function:

$$\mu_f(y_i(k)) = \exp\{-(y_i(k)-u_f)^2/2\sigma^2\}$$

$$\mu_f(y_o(k)) = \exp\{-(y_o(k)-u_f)^2/2\sigma^2\}$$

where $u_f$ is the middle point of the support of each fuzzy set. Since the membership functions are defined, an operator FZ can be defined to transform each crisp value into its corresponding fuzzy membership values of these fuzzy sets in Table 1. Namely,

$$Y_i(k) = FZ(y_i(k)) = \mu_f(y_i(k))$$

$$Y_o(k) = FZ(y_o(k)) = \mu_f(y_o(k))$$

where $y_i(k)$, $y_o(k)$ are crisp values and $Y_i(k)$, $Y_o(k)$ are values representing their corresponding fuzzy membership values.

### B. FUZZY RULE BASE AND COMBINATION OF THE FUZZY CONTROL RULES

A fuzzy rule base consists of fuzzy control rules. There are several ways of deriving fuzzy control rules (Huang 1990, Lee 1990). The fuzzy control rules in this paper are derived from the smoothness assumption. Each fuzzy control rule in the rule base (Table 2) is an If-then rule including two variables $(Y_i(k), Y_o(k))$, for example:

R1: if ( $Y_i(k)$ is pvvb ) then ($Y_o(k)$ is nvvb)

An If-then rule: If P1 then P2 , where P1 represents (X is A) and P2 represents (Y is B), can be defined in many different ways (Lee 1990). For control problems, the proposition P2 is the reaction of the controller when proposition P1 is observed. The If-then rules considered here are only applicable when the proposition P1 is "true". In this situation, the If-then rule is viewed as a cartesian product rule. For example, in the crisp case, the If-then rule: If

Table 2: Fuzzy If-then Rules

| Rule No. | Rules |
|---|---|
| 1 (7) | if $Y_i(k)$ is pvvb (nvvb) then $Y_o(k)$ is nvvb (pvvb) |
| 2 (8) | if $Y_i(k)$ is pvb (nvb) then $Y_o(k)$ is nvb (pvb) |
| 3 (9) | if $Y_i(k)$ is pb (nb) then $Y_o(k)$ is nb (pb) |
| 4 (10) | if $Y_i(k)$ is pm (nm) then $Y_o(k)$ is nm (pm) |
| 5 (11) | if $Y_i(k)$ is ps (ns) then $Y_o(k)$ is ns (ps) |
| 6 (12) | if $Y_i(k)$ is pvs (nvs) then $Y_o(k)$ is nvs (pvs) |
| 13 | if $Y_i(k)$ is ze then $Y_o(k)$ is ze |

X is A, then Y is B is viewed as $I_{A \times B}(x,y) = I_A(x) I_B(y)$ ( or $\min(I_A(x), I_B(y))$ ), where the I is the indicator function. This needs to be extended for a fuzzy If-then rule in this paper. By fuzzy implication (Nguyen 1992), $I_{A \times B}$ is extended to be $\mu_R = \min(\mu_A(x), \mu_B(y))$. Therefore, each of these fuzzy control rules represents one fuzzy subset of the fuzzy set "appropriate control" with the membership function as $\mu_R$, for example,

$$\mu_{R1} = \min(\mu_{pvvb}(Y_i(k)), \mu_{nvvb}(Y_o(k)))$$

where the function "min" is chosen for the fuzzy implication. In order to generate the fuzzy set "appropriate control" (ACL), the fuzzy "or" operator is used to combine these control rules (fuzzy subsets). Symbolically,

$$ACL = R1 \text{ "or" } R2 \text{ "or" } \ldots \text{"or" } Rn$$

with membership function

$$\mu_{ACL}(Y_i(k), Y_o(k)) = \max(\mu_{R1}(Y_i(k) Y_o(k)), \mu_{R2}(Y_i(k), Y_o(k)),$$
$$\ldots, \mu_{Rn}(Y_i(k), Y_o(k)))$$

where the function "max" is the function chosen for the fuzzy "or" operator, $Y_i(k), Y_o(k)$ are functions of $(\theta(k), \theta_{adj}(k))$, respectively, and $\mu_{ACL}$ is a function of $\theta(k)$ and $\theta_{adj}(k)$.

### C. DEFUZZIFICATION

The fuzzy set "appropriate control" is defined with its membership function, $\mu_{ACL}$. For each given input variable, $\theta(k)$, the membership function $\mu_{ACL}$ gives the degree of appropriateness to each value of the output variables $\theta_{adj}(k)$ inside the region $[-\pi/3, \pi/3]$. Defuzzification transforms the fuzzy set "appropriate control" to a nonfuzzy value for the output variable $\theta_{adj}(k)$. The choice of centroid, which is the standard technique in design of a fuzzy controller (Kosko 1992), is used as the defuzzification procedure, that is,

$$\theta_{adj}(k) = (\int \theta_{adj}(k) \mu_{ACL}(\theta(k), \theta_{adj}(k)) d\theta_{adj}(k)) /$$
$$(\int \mu_{ACL}(\theta(k), \theta_{adj}(k)) d\theta_{adj}(k))$$

## IV. SIMULATION RESULTS

Two target trajectory functions (exponential functions) with noise are created for the target position measurements in the simulations. The noise is assumed to be random normally distributed with zero mean and variance less than 0.5. For each target trajectory function, 30 simulations are made with the noise randomly generated for each simulation. The Kalman filter model is represented as

state equation : $x(k+1) = Fx(k) + Gu(k)$

measurement equation : $y(k) = Hx(k) + w(k)$

where $F = G = H = 1$ for one dimension position data. The noise signals $u(k)$ and $w(k)$ are zero mean with variances

$$E\{u(k)u(j)\} = Q\delta(k-j), \quad E\{w(k)w(j)\} = R\delta(k-j)$$

The Kalman filter algorithm here is given by

$$p(k+1|k) = p(k|k) + Q$$

$$K(k+1) = p(k+1|k)(p(k+1|k) + R)^{-1}$$

$$p(k+1|k+1) = (1-K(k+1))p(k+1|k)$$

$$x(k+1) = x(k) + K(k+1)(z(k+1)-x(k))$$

where Q is assumed to be 1, Because the observation noise is generated with variance less than .5, R is assumed to be 0.5.

Concerning the sensitivities in Kosko's work (Kosko 1992), the performance of the filter based on the fuzzy approach is compared to the Kalman filter described above using the sum of the absolute value of the error at each time index k as

$$C_i = \sum_k \left| x_f(k) - x_r(k) \right| \quad i=1,2$$

to be the criterion, where $x_f(k)$ is the filtered target position, $x_r(k)$ is the real target position, $C_1$ represents the simulation result using fuzzy filter, and $C_2$ represents the simulation result using the Kalman filter.

The 30 simulation results show that in about 90% of the simulations the fuzzy filter performs better than the Kalman filter in the sense that $C_1$ is less than $C_2$. Figure 2 gives a comparison of both estimated trajectories when there is no measurement noise. Figure 3 gives $C_1$ and $C_2$ for 30 simulations. Figure 4 and 5 shows the simulation results when noise exists.

## V. CONCLUDING REMARKS

As with the current tendency of exploring expert knowledge in the field of control as a complementary technique to traditional ones when a standard framework is not available, this paper presents a similar case in the area of target tracking. The performance of the fuzzy logic filter in this simple situation is an encouraging sign for further studies in more complex tracking problems.

In the simulations considered, the noise in measurements is modeled by random variables, but the controlling procedure is based on smoothness assumption expressed in natural language which exhibits fuzziness. In other problems, more sophisticated expert knowledge is required. As in AI problems when fuzzy sets are used to model mathematically vague concepts in linguistic information, the choice of logical operations for combining data as well as the determination of membership functions of fuzzy sets involved is still an art which depends on each situation at hand. This flexibility might be beneficial. Empirical results suggest, however, that theoretical explanations are necessary to transform an art into a science. In the literature, the theoretical justification of fuzzy control in Wang and Mendel (Wang 1991) is striking. In this paper, the simulations empirically demonstrate that, "in the long run", the fuzzy filter is superior to the Kalman filter "most of the time", at least for the type of the tracking problems considered here. A more theoretical study is under consideration for future work.

## Acknowledgements

The authors would like to thank all members of the research group on alternative uncertainty models for intelligent systems at New Mexico State University for their friendly discussions and critiques on this work. Special thanks are expressed to Professor H.T. Nguyen for his patience in explaining modern uncertainty techniques.

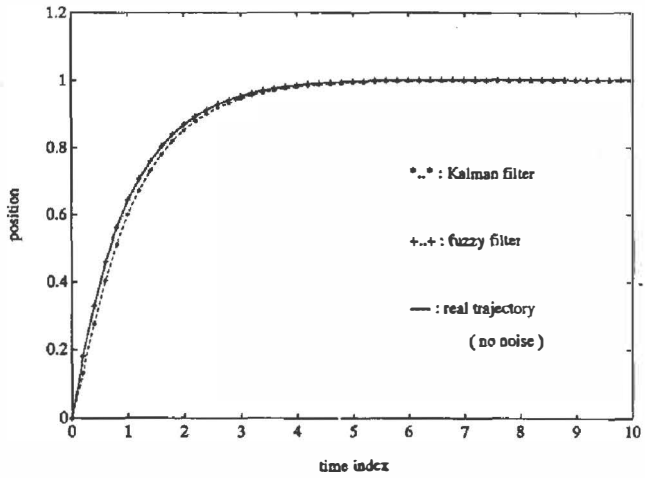

Figure 2: Comparison of Kalman Filter and Fuzzy Filter Tracking.

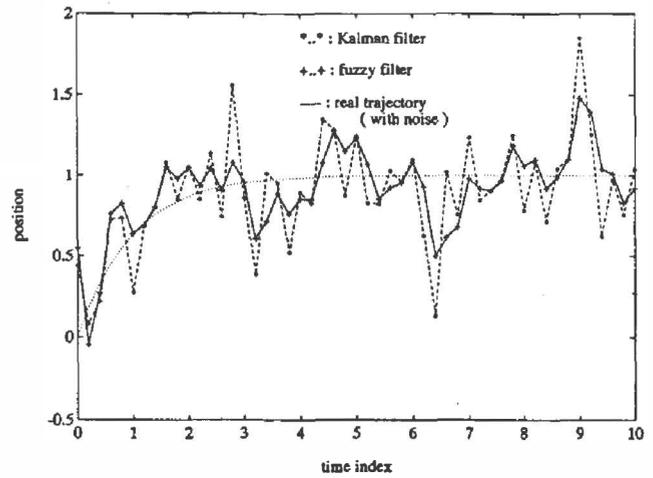

Figure 4: Comparison of Kalman Filter and Fuzzy Filter Tracking

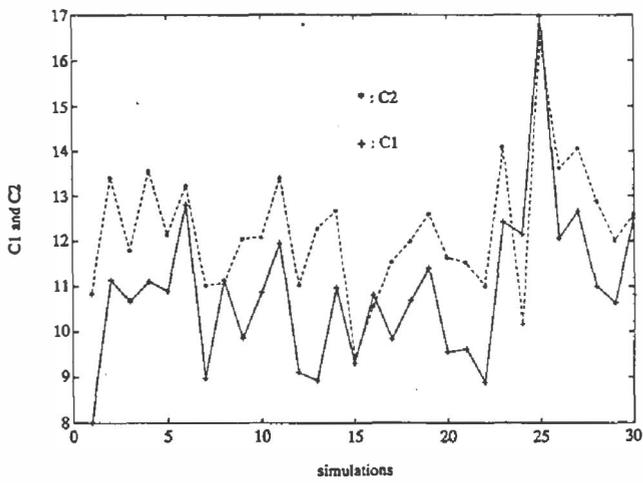

Figure 3: Fuzzy Filter Tracking (C1) and Kalman Filter Tracking (C2) Results for 30 Simulations.

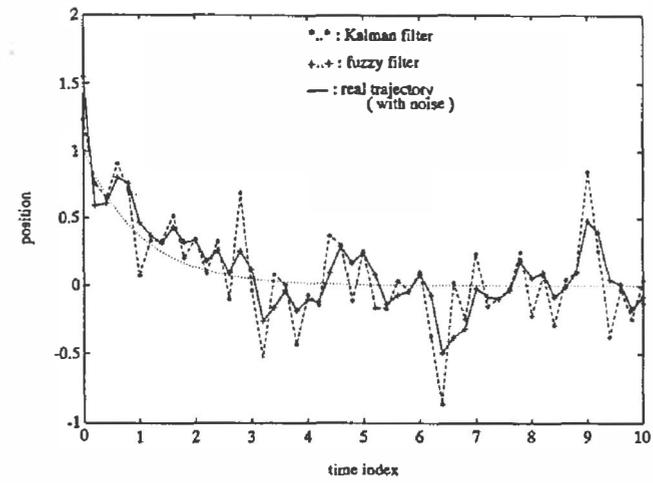

Figure 5: Comparison of Kalman Filter and Fuzzy Filter Tracking